\documentclass[lnicst]{svmultln}
\usepackage{amssymb,amsmath,array}
\usepackage{subfigure}
\usepackage{makeidx}  % allows for indexgeneration
\usepackage[dvips]{graphicx}
\usepackage{psfrag}
\newcommand{\argmin}[2]{\underset{#1}{\mathrm{arg\,min}}\left\{#2\right\}}

\begin{document}

\mainmatter
\title{Bio-inspired speed detection and discrimination}

\titlerunning{bio-inspired motion detection}
\author{Mauricio Cerda\inst{1} \and Lucas Terissi\inst{2} \and Bernard Girau\inst{1}}

\authorrunning{Mauricio Cerda et al.}

\tocauthor{Mauricio Cerda, Lucas Terissi, Bernard Girau}
\institute{Loria - INRIA Nancy Grand Est, Cortex Team\\
Vandoeuvre-l\`es-Nancy - France,\\
\email{[cerdavim,girau]@loria.fr},
\and
Laboratory for System Dynamics and Signal Processing,\\
Universidad Nacional de Rosario - CIFASIS - CONICET - Argentina,\\
\email{terissi@cifasis-conicet.gov.ar}}

\maketitle                % typeset the title of the contribution
% \index{Cerda, Mauricio} % entries for the author index
% \index{Terissi, Lucas}  % of the whole volume
% \index{Girau, Bernard}

\begin{abstract} In the field of computer vision, a crucial task is
the detection of motion (also called optical flow extraction). This
operation allows analysis such as 3D reconstruction, feature tracking,
time-to-collision and novelty detection among others. Most of the
optical flow extraction techniques work within a finite range of
speeds. Usually, the range of detection is extended
towards higher speeds by combining some multi-scale information in
a serial architecture. This serial multi-scale approach suffers
from the problem of error propagation related to the number of scales used in the algorithm. On the other hand, biological experiments show that human
motion perception seems to follow a parallel multi-scale scheme. In
this work we present a bio-inspired parallel architecture to perform
detection of motion, providing a wide range of operation and avoiding
error propagation associated with the serial architecture. To test our
algorithm, we perform relative error comparisons between both classical and
proposed techniques, showing that the parallel architecture is able to achieve
motion detection with results similar to the serial approach.

\end{abstract}

\section{Introduction} The visual capabilities in humans have
motivated a large number of scientific studies.
The performance to perceive and interpret visual stimuli in
different species including humans is outstanding: the wide range of
tolerance to different illumination and noise levels are just a few
characteristics that we are aware of, but that are still barely
understood.

An important aspect in visual processing is the perception of
motion. Motion is a key step in several computer vision tasks such as
3D reconstruction, feature tracking, time-to-collision estimation,
novelty detection, among others \cite{key:barron}. Motion is also one
of the features that many species can perceive from the flow of visual
information, and its detection has been observed in a large number of
animals \cite{key:visualneuro}, from invertebrates to highly evolved
mammals. From optical engineering and experimental psychology we
already know the main features of human motion
discrimination \cite{key:nakayama}.
In this work, we are particularly interested in taking
inspiration from biology in order to design a parallel algorithm
with similar discrimination capabilities as obtained by classical serial
architectures.

Our work begins by presenting an overview of techniques to detect
motion in machine vision to continue with the available experimental
results and their procedures in human psychophysics. Section 3,
presents our algorithm for speed detection with which we perform our
simulations, and compare with experimental data. The results are
analyzed in section 4, and in the last two sections, we present the
discussion and conclusions about our work.

\section{Overview}

In this work we are interested in the detection of motion,
specifically in the coding and retrieval of speed ($\vec{v}$), and in
the link between the idea of selecting a range of speed to work with,
and providing wider ranges of discrimination as observed in human
psychophysics experiments \cite{key:orban}. We focus on two features:
the multi-scale architecture of the speed detection, and the relation between the number of multi-scale levels and the range of speeds the system is sensitive to.

\subsection{Motion detection in computer vision}\label{subsec:cv}
The detection of motion is a widely used operation in computer
vision. Commonly called ``optical flow extraction'', the main
objective is to assign a vector $\vec{v}=\left(u,v\right)$ to each
frame pixel from a given sequence of frames (at least
two). In this section, we explain the
basic technique to increase the motion range of an optical flow
extraction which the method is sensitive to. We ground our explanation on
the well-known Lucas \& Kanade's
method \cite{key:lucas,key:barron} (the basic multi-scale technique
similarly applies to other methods for optical flow extraction).

\subsubsection{Optical flow}
Many optical flow extraction methods are based on the initial assumption of brightness conservation, that is,
\begin{equation}
\frac{dI(x,y,t)}{dt}=\frac{\partial I}{\partial x}u + \frac{\partial I}{\partial y}v + \frac{\partial I}{\partial t}=0
\label{eq:bc}
\end{equation}

\noindent where $\vec{v}=(u,v)$ is the velocity vector. A well known technique
following this approach is the Lucas \& Kanade algorithm \cite{key:lucas}, that minimizes the following cost function in a small fixed region $\Omega$, \emph{i.e.}

\begin{equation}
\vec{v}=\argmin{\vec{v}}{
\sum_{\vec{x}\in\Omega} W^2(\vec{x})\left[\nabla
I(\vec{x}, t)\cdotp \vec{v} + \frac{\partial I}{\partial t}(\vec{x}, t)\right]^2}
\label{eq:lucaskanade1}
\end{equation}

\noindent where $W$ is a two-dimensional Gaussian function used
to give more importance to the central points and $\Omega$ is a square
region of a few pixels. This minimization estimates $\vec{v}$ with
sub-pixel precision after a few iterations. This method achieves good
optical flow extraction
in regions where $\left|\nabla I(\vec{x}, t)\right| > 0$,
such as corners \cite{key:simon}.

\subsubsection{Serial multi-scale optical flow}
The Lucas \& Kanade method for optical flow extraction
considers a small region $\Omega$. The use of this region $\Omega$ is not
particular to this method: it is used in most
algorithms~\cite{key:barron}. As the computation is performed in small
windows, the detection of motion is constrained to detect speeds up to
$\omega$ pixels per frame, where $\omega$ stand for the diameter of
$\Omega$. To overcome this limitation, a multi-scale
representation of the images can be performed, usually by considering
Gaussian pyramids~\cite{key:black}. A Gaussian pyramid representation
of an image is computed by recursively smoothing (using a Gaussian
kernel) and sub-sampling the original image. In this way, the original
image is represented by a set of smaller images. The representation at scale
level $l = 0$ is the original image itself. The image at level $l$ is
obtained by sub-sampling a filtered version of the image at level
$l-1$ with a downsampling factor equal to 2. Thus, the size of the
image at each level $l$ is $N_l=N_{l-1}/2$ with $l=1,2,\dots,(L-1)$, where
$L$ is the number of levels of the representation.

In the serial multi-scale optical flow estimation, speed is
computed by sequentially projecting the estimation obtained at level
$l$ to level $l-1$, until level $l=0$. There are complex strategies
for computing the optical flow with a multi-scale
approach~\cite{key:simon}. A simple solution for optical flow
computation is implemented in the widely used computer vision library
OpenCV~\cite{key:opencv,key:black}. In this case, the multi-scale
estimation starts from the highest level ($l=L-1$) and it propagates to
the next one:
\begin{equation} \vec{v}_{l-1}=2*\vec{v}_{l} + d_{l-1}(\vec{v}_{l})
    \label{eq:multi}
\end{equation} where $d_{l-1}$ is the estimation of velocity at
level $l-1$ after projecting the estimation $v_{l}$ by warping the
image by $-v_{l}$ at level $l-1$. Computing the optical flow from
the highest level and then projecting the solution to the lower level
\cite{key:simon,key:black} increases the range of detectable
speeds. This range is wider when more scales are used. On the other
hand, the sequential projection between levels also propagates the
error introduced at each level. Thus, in terms of precision,
increasing the number of scales in the representation increases the
error introduced in the estimation.

\subsection{Biological elements}This section sketches out the current experimental
knowledge in biology, focusing on studies of speed coding in
the human brain \cite{key:logmt} and on higher level descriptions of
speed discrimination from experimental psychophysics
\cite{key:nakayama,key:metha,key:koenderink}.

\subsubsection{Parallel architecture}\label{subsec:architecture} In
the human brain the main area that is responsible for coding different
speeds is area MT \cite{key:visualneuro}. It is located in the
occipital region (back of the head). Neurons in this area are
selective to stimuli moving at a given speed \cite{key:mt}.
Their spatial organization is retinotopical \cite{JingLiu01012003}:
each neuron has a reduced visual field,
and neurons who share the same local visual field are grouped together
in some macro-column that contains cortical columns that
are selective for different orientations. This configuration allows
a complete mapping of the visual field with a group of cortical columns
that codes for all possible directions of local motions.
The spatial organization is less known with respect to the speed
selectivity. Nevertheless, it has been found that (1) the average detected speed increases with eccentricity (with respect to the retinotopical organization of MT), (2) similar speeds are detected by neurons
closer than for distinct speeds, and (3) for each eccentricity, there are neurons for different speeds \cite{JingLiu01012003}.
The interactions between different units is not completely understood, but there is evidence that units sensitive to different speeds could be coding a range of speeds in parallel \cite{key:logmt}. It has been observed that the range of detectable speeds is not uniformly covered \cite{key:mt}, but in this work we are interested in the simultaneous existence of speed selective units in MT that could be accountable for a parallel architecture dealing with different speeds.

\subsubsection{Speed discrimination}\label{subsec:speed_discrimination} In the work of McKee et al. \cite{key:nakayama}, two subjects were exposed to several
stimuli, one of those being a horizontal scaled single bar vertically
moving at different eccentricities. The goal of this experiment was to
determine the minimal relative detectable variation in speed for every
subject with the sight fixed at a certain location and for each stimuli eccentricity.

%\begin{figure}[htp] %\centering
%\includegraphics[width=0.5\textwidth]{figures/experiment.eps}
%\caption{An schematic diagram of the experiment proposed by McKee et
%al. \cite{key:nakayama}. The subject has the sight fix at the ``+''
%and the bar moves at different speeds.}
%\label{fig:setup}
%\end{figure}

It is important to mention how this was actually measured, because the
subject cannot assign a precise velocity at each location. Instead,
given a reference velocity, the subject was asked to indicate whether the
next presented stimuli moves faster or slower. The minimal detectable
variation was then statistically inferred. Related experiments were
performed by others \cite{key:orban,key:metha,key:koenderink},
showing that the measurements are not affected by different contrast
conditions, and that they do not depend on binocular or monocular
sight.

\begin{figure}[htp]
\begin{center}
      \psfrag{v}[][]{\scriptsize $v$ [degrees/sec]}%
      \psfrag{disc}[][]{\scriptsize $\frac{\Delta v}{v}$}%
\includegraphics[width=0.7\columnwidth]{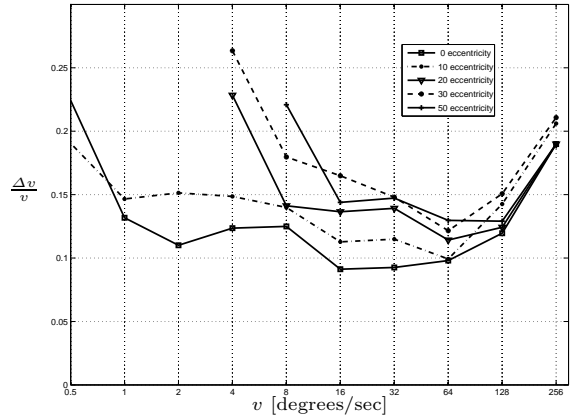}
\caption{Weber fraction (minimum threshold of perceived change in speed or $\Delta v / v$) as function of velocity, for
different eccentricities in the human subject B.D.B. (condition 2), in the
differential motion experiment by \cite{key:orban} (reproduced from their work), each curve is at eccentricities 0,10,20,30 and 50, respectively (left to right). The speed axe is in logarithmic scale.}
\label{fig:result}
\end{center}
\end{figure}

The described experiments study the speed discrimination at several
eccentricities\footnote{Distance to the center of the eye in foveated visition (humans, primates and others).}, see Fig. \ref{fig:result}. In this work we are
interested in each one of these eccentricities and their related
discrimination properties, and not in the relations between different
eccentricities. In order to model these discrimination functions, we need to generate a given discrimination percentage in a range of speed $[v_1,v_2]$. We also point out that the left side of the experimental curves, see Fig. \ref{fig:result}, is related to the eccentricity but the same idea holds: for each eccentricity there is a wide range of speed discrimination, where the relative error (rather than the absolute error) remains stable (5\%-15\%).

%%%%%%%%%%%%%%%%%%%%%%%%%%%%%%%%%%%%%%%%%%%%%%%%%%%%%%%%%%%%%
\section{Proposed parallel multi-scale speed detection}

Multi-scale speed detection is based on the fact that a particular
speed detection algorithm can be used to estimate slower speeds at
lower levels and to estimate faster speeds at higher levels. This
information is used in the above described serial multi-scale optical
flow algorithm to detect speeds in a wide range of velocities by
projecting the information at level $l+1$ to estimate speed at level
$l$, \emph{i.e.} in a serial manner. As it is described in
\cite{key:mt,key:logmt}, it seems that human motion perception is
based on a parallel multi-scale scheme. Based on this idea, the speed
detection algorithm proposed in this paper estimates the speed by
combining the information computed at each level independently,
\emph{i.e.} using the multi-scale information in a parallel
manner. In this case, there is no error propagation on the computation
of speeds at each level because it does not depend on the estimation
performed for other levels. At each level $l$, we compute speeds using the
optical flow estimation algorithm described above, see
subsection~\ref{subsec:cv}. As explained before, this choice
does not bias our results, since our work is to provide a
bio-inspired parallel speed detection instead of the standard serial
approach, for any optical flow extraction method.

\begin{figure}
  \begin{center}
      \psfrag{l=1}[][]{\scriptsize ~~~$l=0$}%
      \psfrag{l=2}[][]{\scriptsize ~~$l=1$}%
      \psfrag{l=3}[][]{\scriptsize ~~$l=2$}%
      \psfrag{v}[][]{\scriptsize $v_i$ [pixels/frame]}%
      \psfrag{k(v)}[][]{\scriptsize $k_l(v_i)$}%
      \psfrag{k_est(v)}[][]{\scriptsize $\hat{k}_l(v_i)$}%
      \subfigure[]{
      \includegraphics[width=0.47\columnwidth]{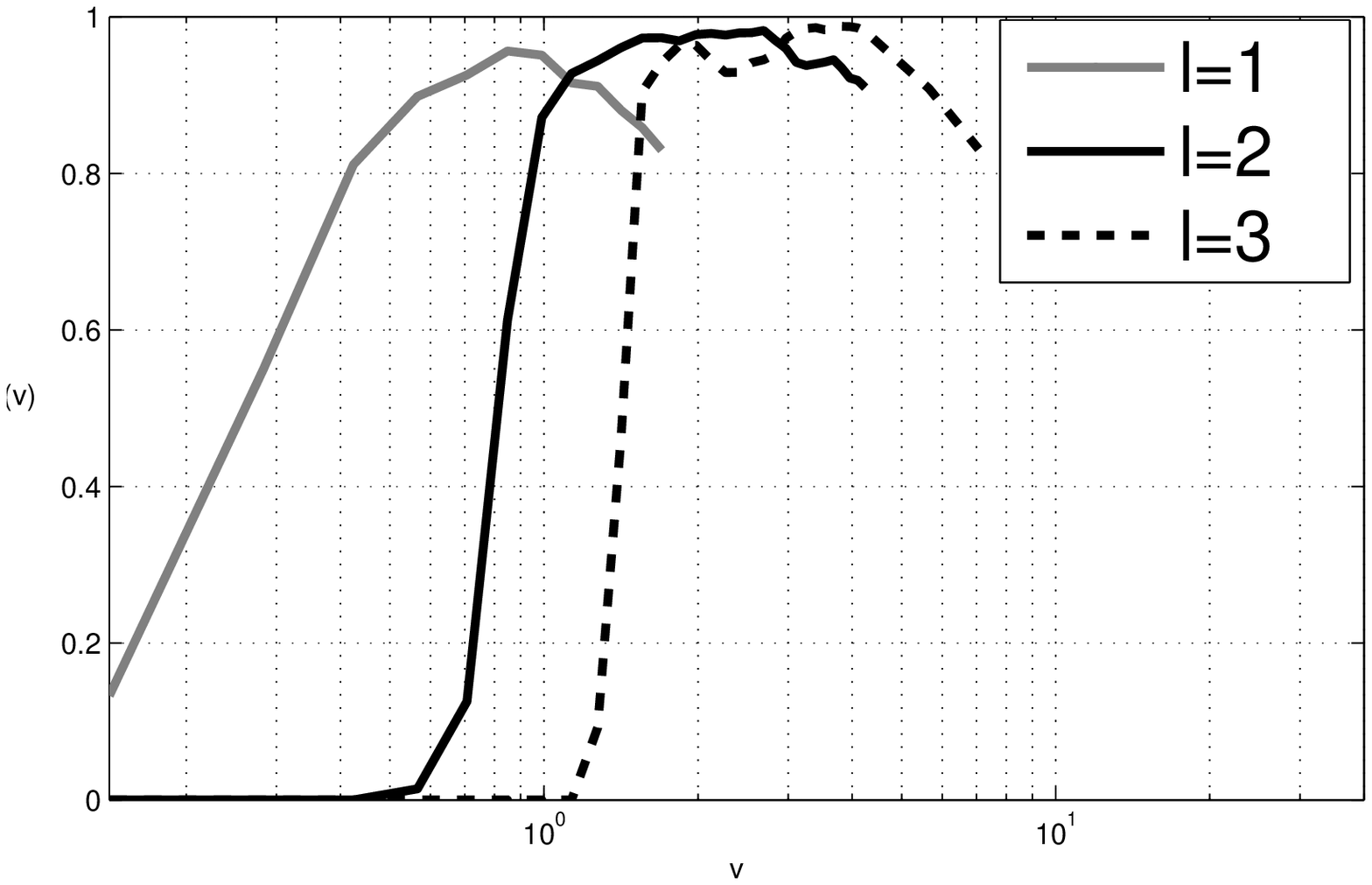}
      \label{fig:confidence_OF_n123}}
      \subfigure[]{
      \includegraphics[width=0.47\columnwidth]{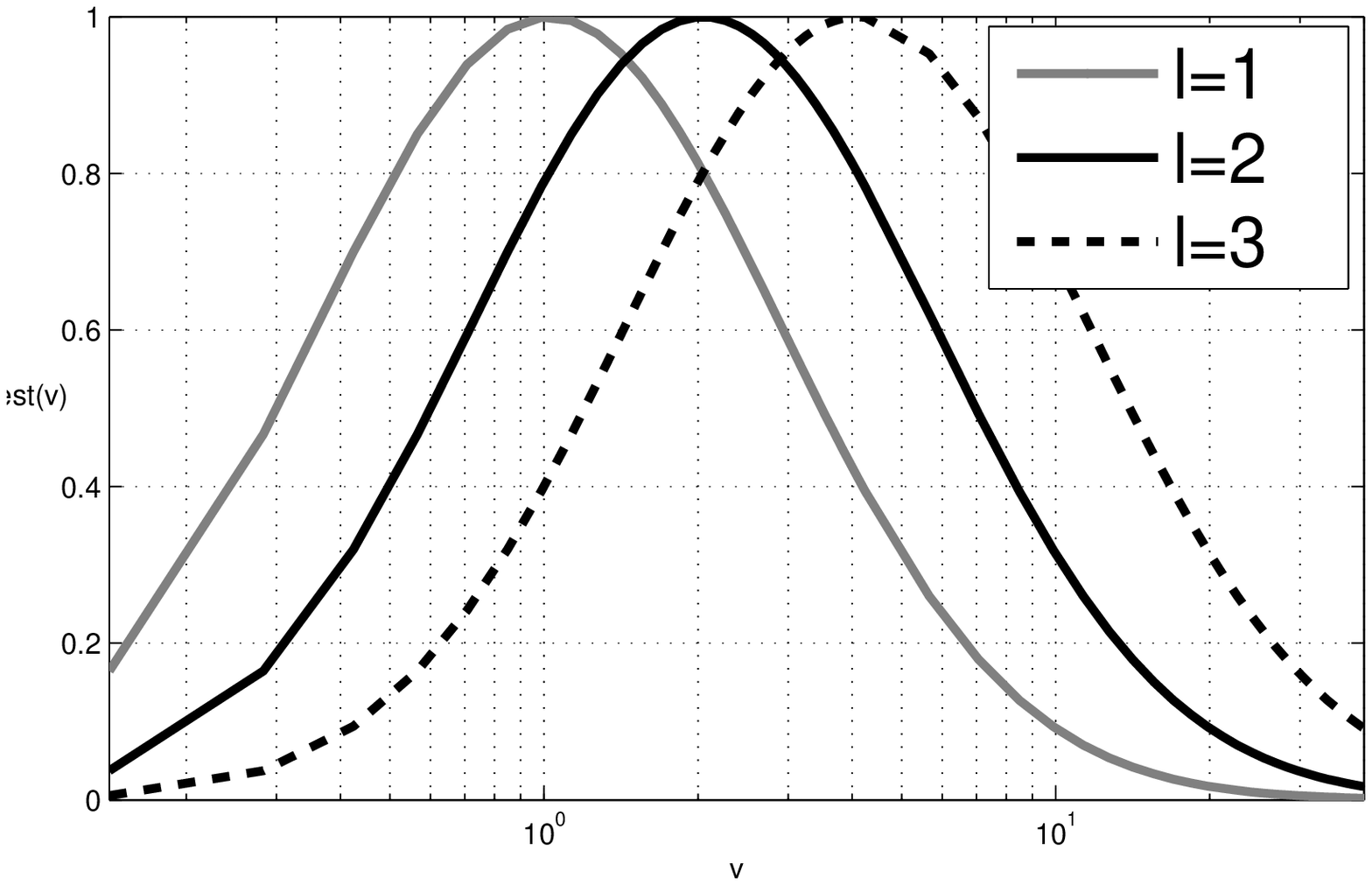}
      \label{fig:confidence_OF_n123_est}}
      \caption{Confidence distribution $k_l$ for different levels $l$. (a) Experimental distributions~$k_l$. (b) Approximated distributions $\hat{k}_l$.}
  \end{center}
\end{figure}

\begin{figure}
  \begin{center}
      \psfrag{Frame 0}[][]{\scriptsize Frame 0}%
      \psfrag{Frame 1}[][]{\scriptsize Frame 1}%
      \psfrag{Optical Flow}[][]{}%
      \subfigure[]{
      \includegraphics[width=0.46\columnwidth]{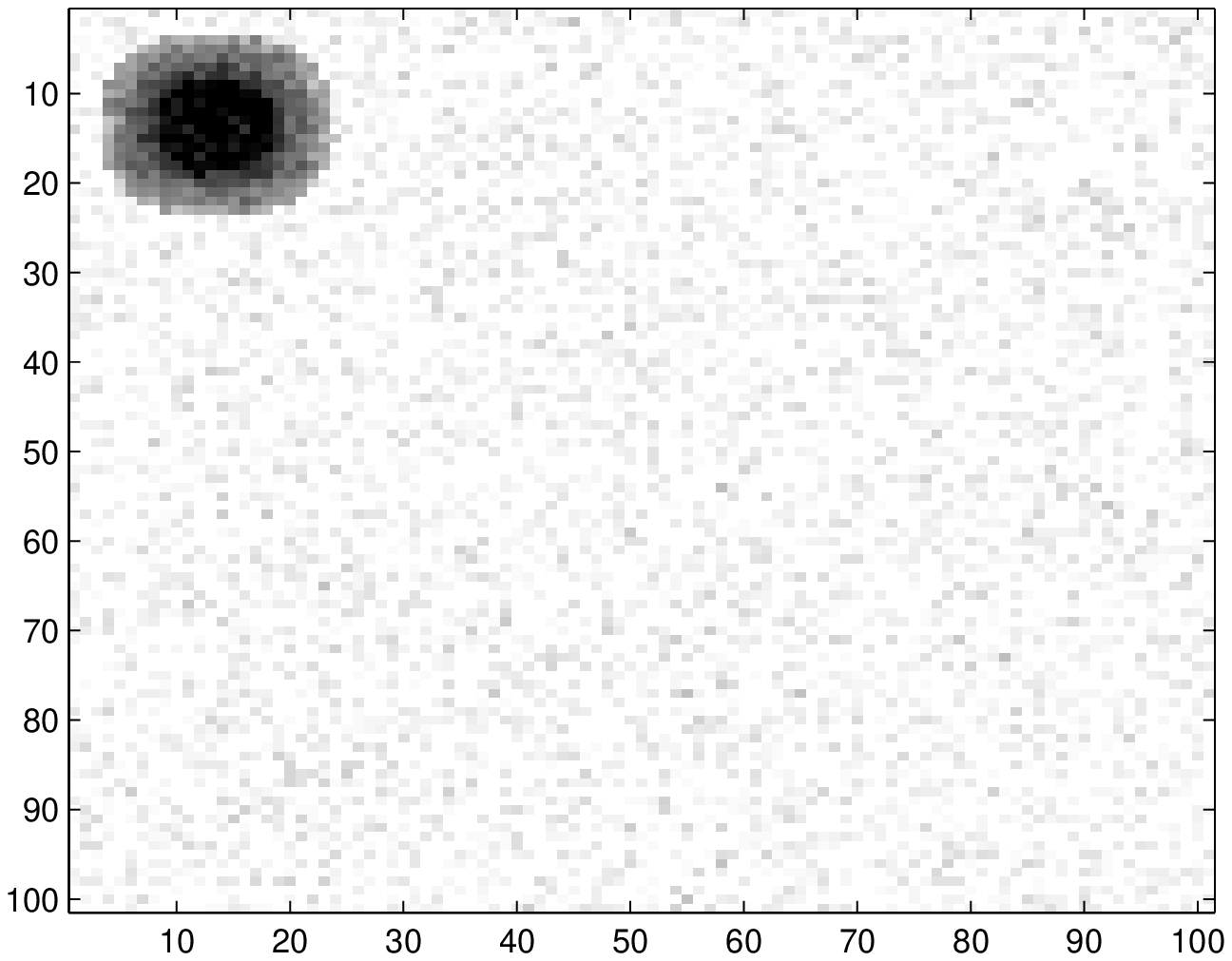}
      \label{fig:input1}}
      \subfigure[]{
      \includegraphics[width=0.46\columnwidth]{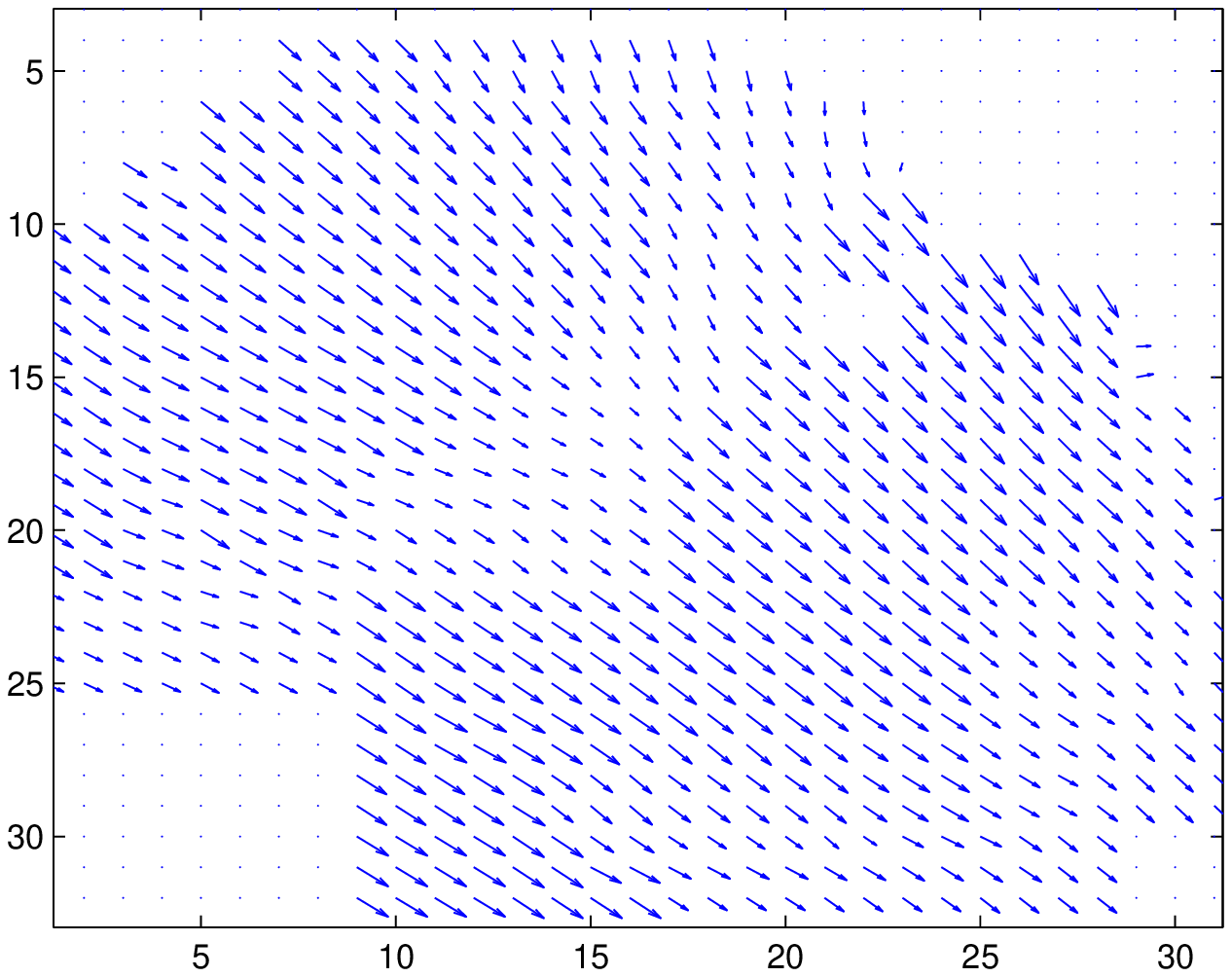}
      \label{fig:input3}}
      \caption{In (a) one frame of an input image sequence used in the experiments is depicted ($v=10$ bottom-right direction). (b) is the obtained optical flow using the proposed parallel multi-scale algorithm. Note that optical flow image was zoomed around the object position.}
  \end{center}
\end{figure}

As expected, the speed detection algorithm estimates speed with
a certain error at each multi-scale level $l$. The confidence in the
estimation of speed $v_r$ at level $l$, denoted as $k_l(v_r)$, can be
defined as
\begin{equation}\label{eq:confidence}
k_l(v_r) = 1 - \left|\frac{v_r-v_e}{v_r} \right|
\end{equation}
where $v_r$ is the magnitude of the object's real speed
($v_r=\|\vec{v}_r\|$) and $v_e$ is the magnitude of the average estimated
speed on the object pixels location. It can be noted, that this
computation only takes into account the magnitude of the speed,
ignoring its direction. Figure~\ref{fig:confidence_OF_n123} shows the
confidence $k_l(v_r)$ for three different multi-scale levels. These
distributions were computed using an input image sequence containing
an object moving at different speeds in a range from 0.5 pixels per
frame to 20 pixels per frame. To statistically determine the
confidence at each level $l$ and speed $v_r$, the experiments were
carried out using the input image sequence with several realizations
of Gaussian white noise, then the resulting confidence $k_l(v_r)$ is
computed as the mean value of the ones obtained in the
experiments. Figure~\ref{fig:input1} shows two
frames of an input image sequence used in the experiments. In this
sequence the object is moving at 10 frames per pixel in the bottom-right
direction.

As it may be seen in Figure~\ref{fig:confidence_OF_n123}, a particular
speed $v_r$ can be detected at several multi-scale levels but with
different confidence values. Thus, the current speed could be
estimated by taking into account the speeds computed at each level $l$
and their associated confidence values $k_l$. For that reason, the
experimental distributions depicted in
Fig.~\ref{fig:confidence_OF_n123} have to be approximated by a
closed-form equation. In this work, these distributions are
approximated (modeled) as Gaussian distributions in a semi-log space
defined by the following equation
\begin{eqnarray}\label{eq:confidence_aprox}
\hat{k}_l(v_r) &=& \mathrm{exp}\left(-\left[\frac{\mathrm{log}(v_r) -
\mu_l}{\sigma_l}\right]^2 \right)\\ \mu_l &=& \mu_0 +
\mathrm{log}(c^{l-1})\\ \sigma_l &=& \sigma_0
\end{eqnarray}
where $\mu_0$ and $\sigma_0$ are the mean and variance
of the distribution at level $l=0$ and $c$ is the scaling factor used
in the sub-sampling of the images. The approximated distributions
$\hat{k}_l$ for $l=0$, $l=1$ and $l=2$ are depicted in
Fig.~\ref{fig:confidence_OF_n123_est}. It may be seen that a better
approximation of the distributions could be obtained using a
particular set of variables for each level but this would increase the model
complexity. The approximation of the distributions $\hat{k}_l$ for
each level in Eq.~(\ref{eq:confidence_aprox}) only depends on
$\mu_0$, $\sigma_0$ and $c$. It may be noted that this approximation
allows to perform the estimation of speeds using different values of the scaling factor $c$, which is usually set to $c=2$, \emph{i.e.}, the case of using Gaussian pyramids for the sub-sampling.

\begin{figure}
  \begin{center}
      \psfrag{l=1}[][]{\scriptsize ~~~~~$l=0$}%
      \psfrag{l=2}[][]{\scriptsize ~~~~$l=1$}%
      \psfrag{l=3}[][]{\scriptsize ~~~~$l=2$}%
      \psfrag{v}[][]{\scriptsize $v_i$ [pixels/frame]}%
      \psfrag{k(v)}[][]{\scriptsize $k_l(v_i)$}%
      %\psfrag{proposed l=3 c=2}[][]{\scriptsize parallel, $L=3$, $c=2$}%
      \psfrag{proposed l=3 c=2}[][]{\scriptsize parallel $L=3$}%
      \includegraphics[width=0.6\columnwidth]{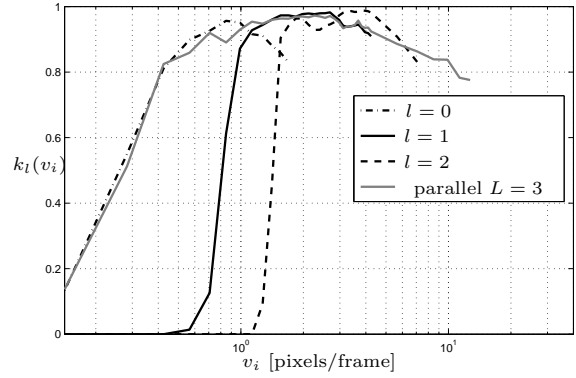}
      \caption{Comparison between confidence distribution of the proposed algorithm (parallel) with $L=3$ and $c=2$, and confidence distributions $k_l$ for levels $l=0,1$ and 2 without projection between them.}
      \label{fig:confidence_comparison}
  \end{center}
\end{figure}

Finally, denoting the detected speed at each level
$l$ by $\vec{v}_e^l$, the proposed algorithm computes the current speed, using the speed detected at each multi-scale level with its associated
confidence value $\hat{k}_l(\|\vec{v}_e^l\|)$, as
\begin{equation}\label{eq:parallel_final_speed}
\vec{v}_f =
\frac{\sum_{l=0}^{L-1} \vec{v}_e^l
\hat{k}_l(\|\vec{v}_e^l\|)}{\sum_{l=0}^{L-1} \hat{k}_l(\|\vec{v}_e^l\|)}
\end{equation}
where $L$ is the number of levels used to compute the
estimated speed $\vec{v}_f$. Figure~\ref{fig:input3} shows the obtained optical flow using the proposed parallel multi-scale algorithm.
The comparison between the experimental confidence
distribution of the proposed algorithm and confidence distributions
for three levels is shown in Fig.~\ref{fig:confidence_comparison}. As
expected, the confidence distribution of the parallel
multi-scale algorithm with $L=3$ is approximately the envelope of
the confidence distributions of levels $l=0$, $l=1$ and $l=2$.

%\begin{figure}
%  \begin{center}
%      \psfrag{L=1}[][]{\scriptsize ~~~$L=1$}%
%      \psfrag{L=2}[][]{\scriptsize ~~$L=2$}%
%      \psfrag{L=3}[][]{\scriptsize ~~$L=3$}%
%      \psfrag{v}[][]{\scriptsize $v_i$ [pixels/frame]}%
%      \psfrag{D(v)}[][]{$\frac{\Delta v}{v}\%$}%
%      \includegraphics[width=0.6\columnwidth]{figures/discrim_OFC2_123_bn.eps}
%      \caption{Discrimination of the proposed algorithm ($c=2$) for $L=1$, $L=2$ and $L=3$.}
%      \label{fig:discrim_OFC2_123}
%  \end{center}
%\end{figure}

\begin{figure}
  \begin{center}
      \psfrag{L=1}[][]{\scriptsize ~~~~~$L=1$}%
      \psfrag{L=2}[][]{\scriptsize ~~~~$L=2$}%
      \psfrag{L=3}[][]{\scriptsize ~~~~$L=3$}%
      \psfrag{parallel L=2}[][]{\scriptsize ~~~~~~~~~~parallel, $L=2$}%
      \psfrag{serial L=2}[][]{\scriptsize ~~~~~~~~~serial, $L=2$}%
      \psfrag{parallel L=3}[][]{\scriptsize ~~~~~~~~~~parallel, $L=3$}%
      \psfrag{serial L=3}[][]{\scriptsize ~~~~~~~~~serial, $L=3$}%
      \psfrag{parallel L=4}[][]{\scriptsize ~~~~~~~~~~parallel, $L=4$}%
      \psfrag{serial L=4}[][]{\scriptsize ~~~~~~~~~serial, $L=4$}%
      \psfrag{v}[][]{\scriptsize $v_i$ [pixels/frame]}%
      \psfrag{D(v)}[][]{$\frac{\Delta v}{v}\%$}%
      \subfigure[]{
      \includegraphics[width=0.47\columnwidth]{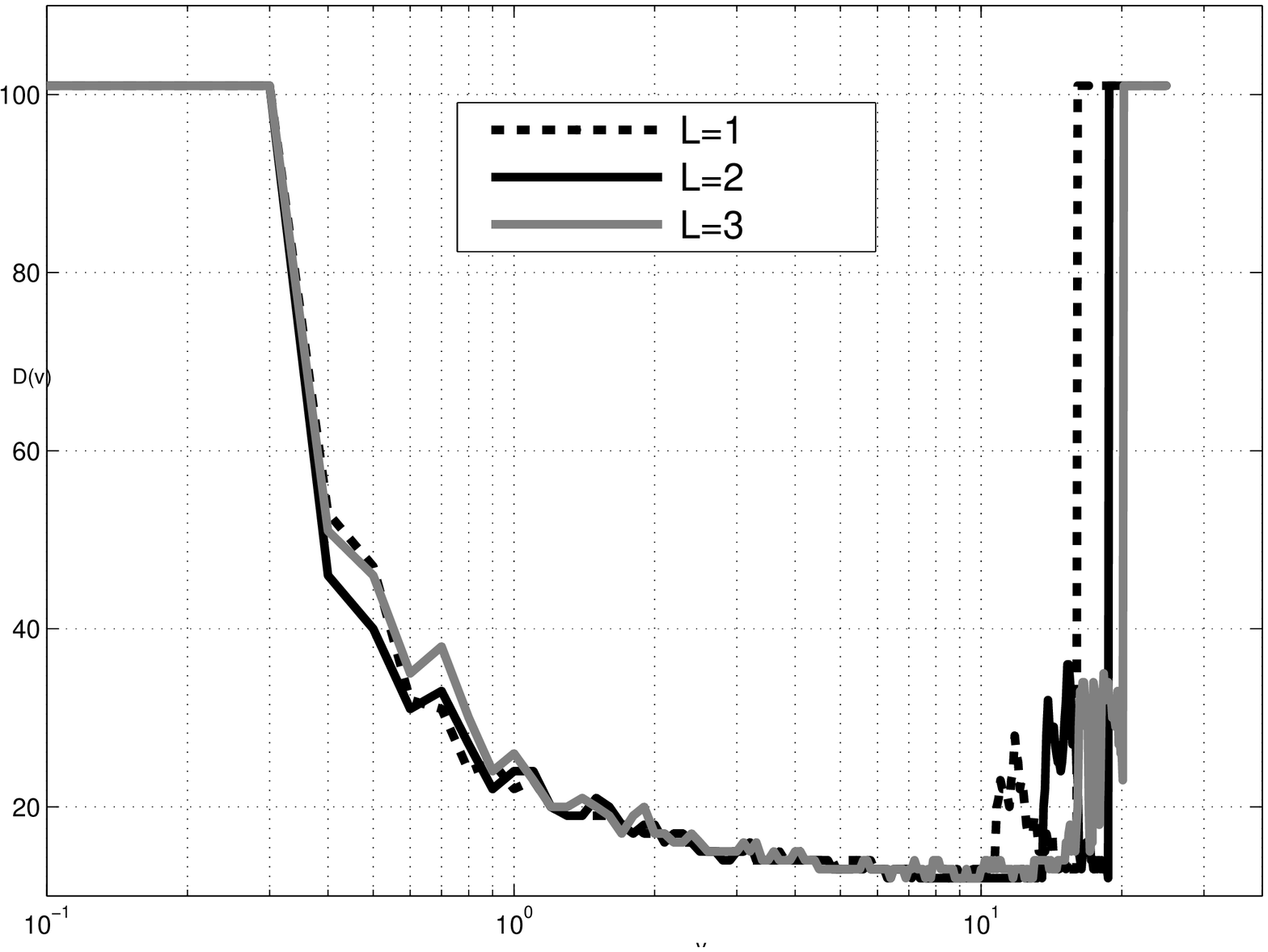}
      \label{fig:discrim_OFC2_123}}
      \subfigure[]{
      \includegraphics[width=0.47\columnwidth]{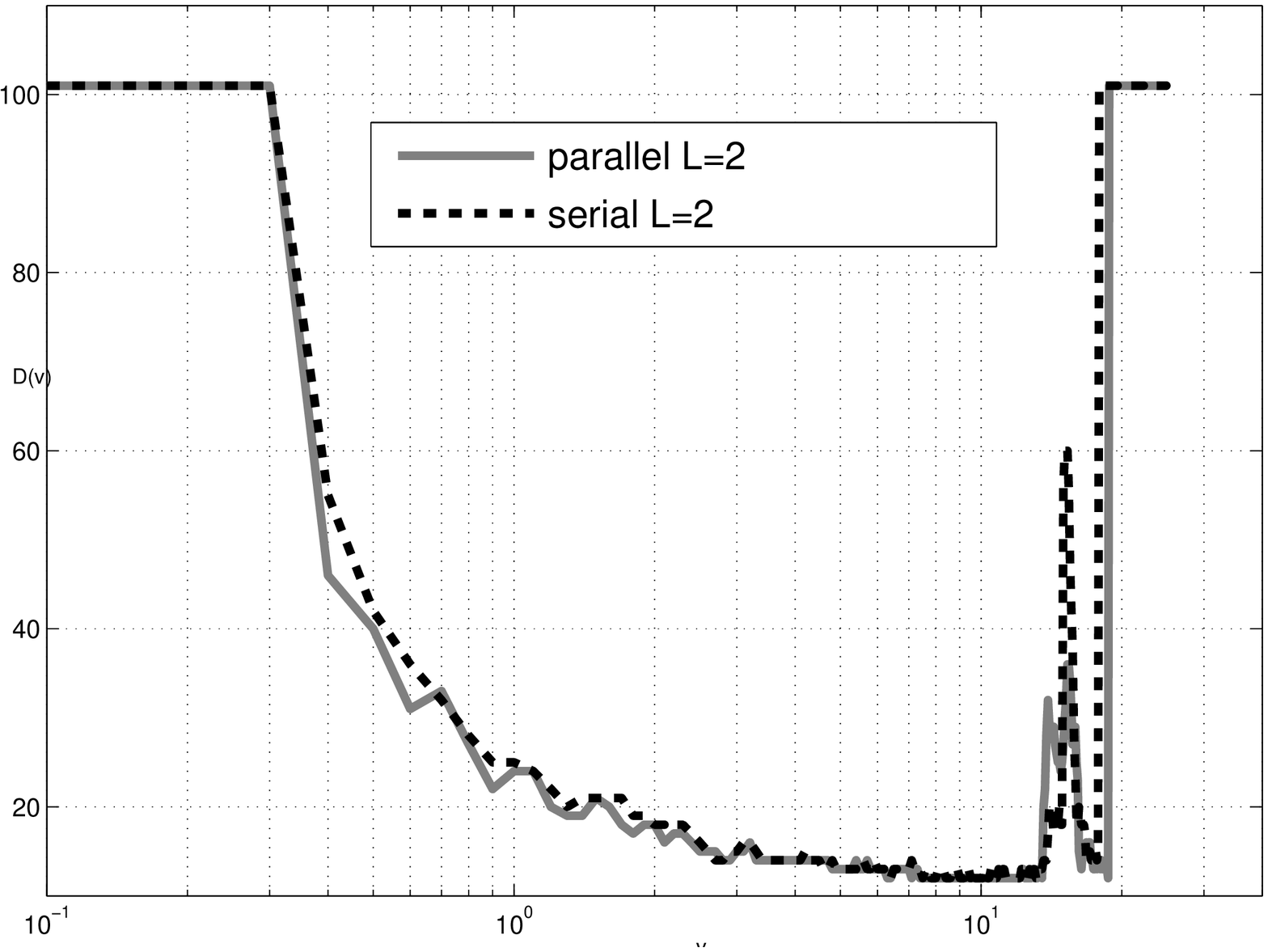}
      \label{fig:discrim_OFC2_OForig_n2}}\\
      \subfigure[]{
      \includegraphics[width=0.47\columnwidth]{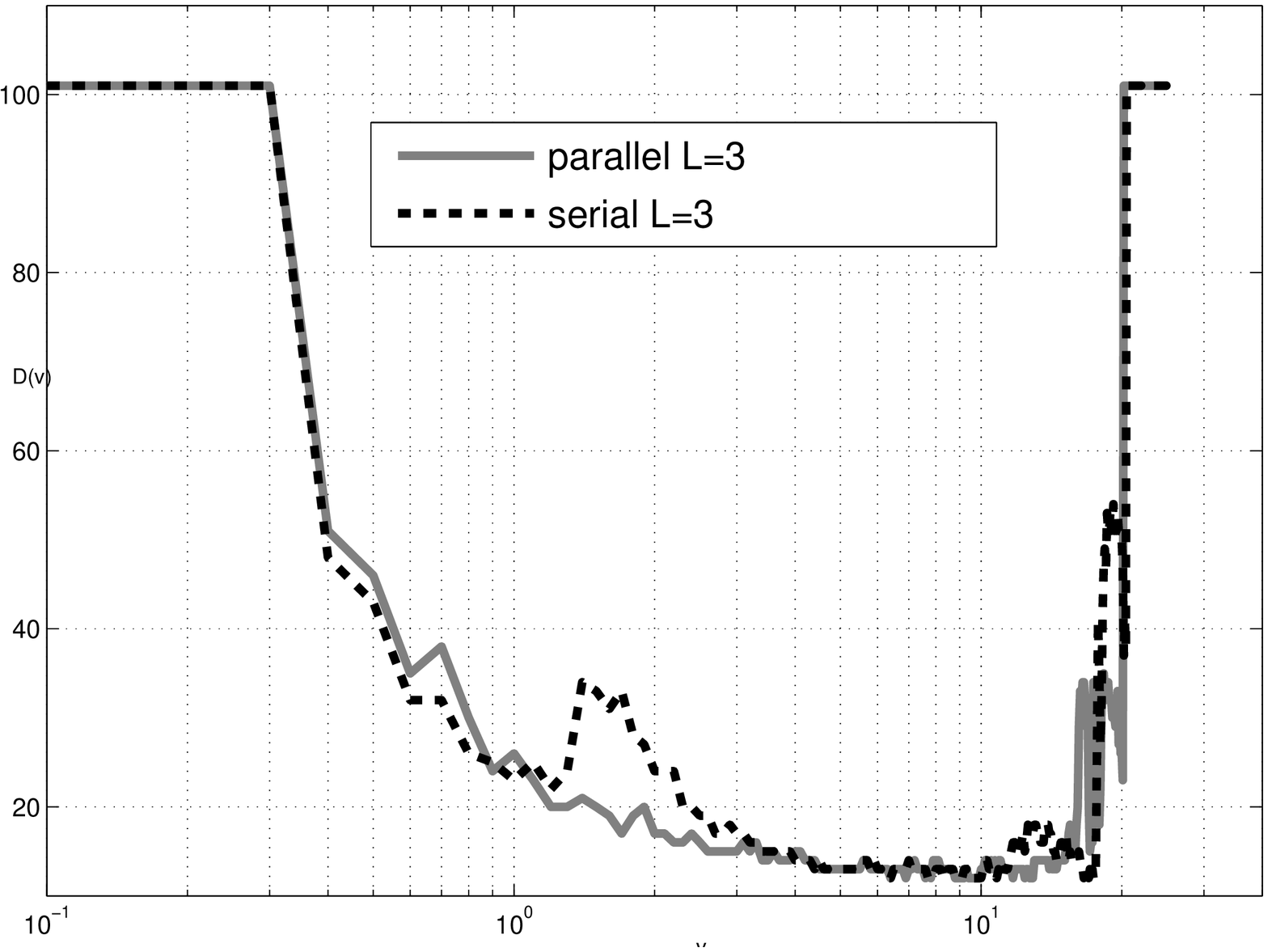}
      \label{fig:discrim_OFC2_OForig_n3}}
      \subfigure[]{
      \includegraphics[width=0.47\columnwidth]{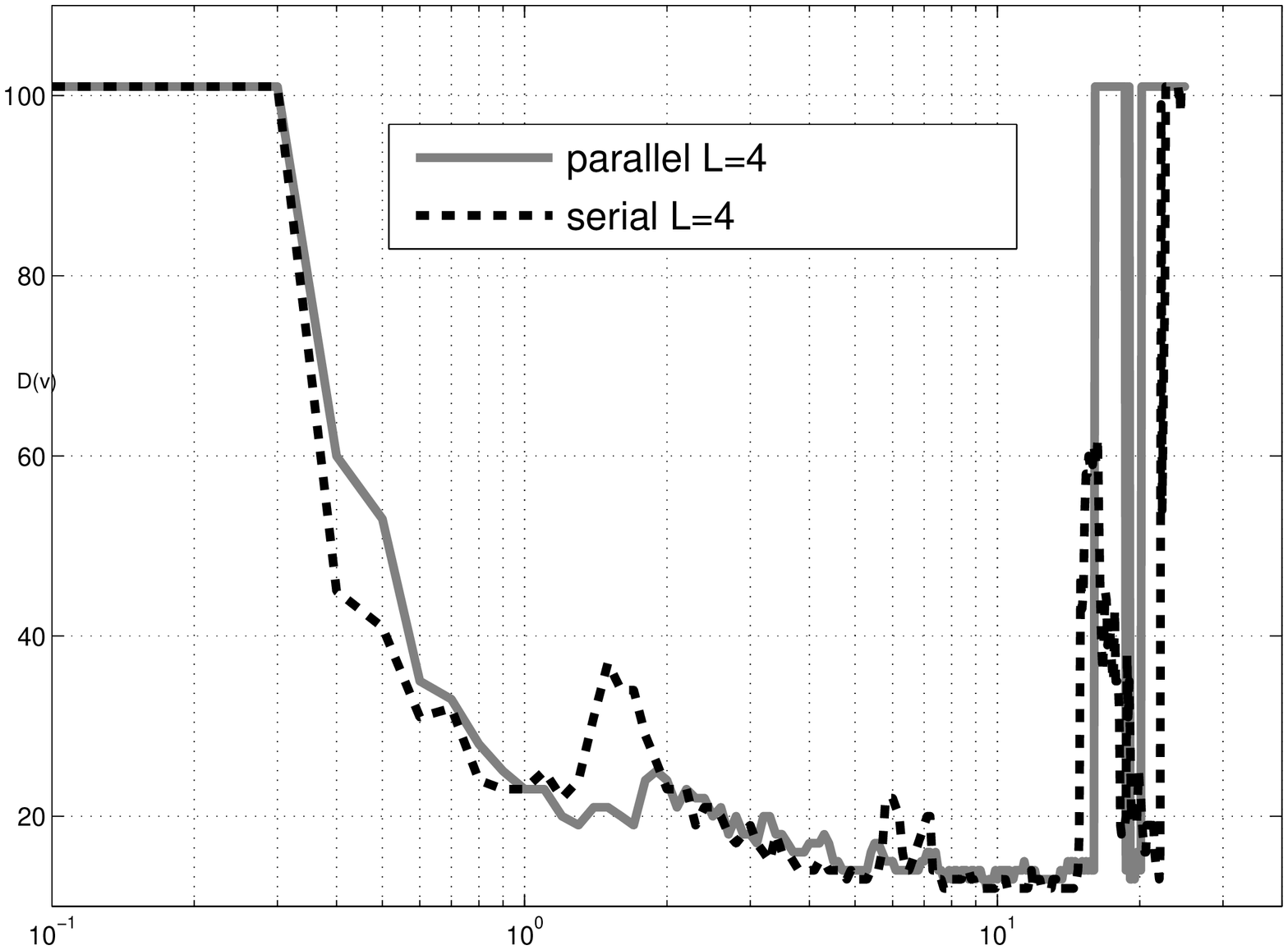}
      \label{fig:discrim_OFC2_OForig_n4}}
      \caption{(a) Discrimination of the proposed algorithm for $L=1$, $L=2$ and $L=3$ ($c=2$). In (b), (c) and (d), the comparisons between the discrimination of the proposed parallel ($c=2$) and the serial algorithms for $L=2$, $L=3$ and $L=4$, respectively, are shown.}\label{fig:discrim_OFC2_OForig}
  \end{center}
\end{figure}

\section{Results}
%method
As it was described in subsection~\ref{subsec:speed_discrimination},
speed discrimination is computed as the minimal detectable variation
in speed of a particular visual stimuli. In this work, a variation in
speed, from a given reference speed $v_{obj}$, of the moving object is
considered to be noticeable if the following inequality holds
\begin{equation}\label{eq:delta_detectable}
\left|\frac{\widehat{v}_{v_{obj}} - \widehat{v}_{v_{obj} \pm \Delta v_{obj}}}{v_{obj}} \right| > \alpha
\end{equation}
where $\widehat{v}_{v_{obj}}$ and $\widehat{v}_{v_{obj} \pm \Delta v_{obj}}$
are the speeds estimated by the algorithm when the object is moving at
velocities $v_{obj}$ and $v_{obj}\pm\Delta v_{obj}$, respectively, and
$\alpha$ is the percentage of variation from the object speed
$v_{obj}$ required to consider $\Delta v_{obj}$ as detectable. It may
be noted in Eq.~(\ref{eq:delta_detectable}) that a variation in speed
is considered noticeable if it is detectable when $v_{obj}$ is both
increased and decreased in $\Delta v_{obj}$. To statistically
determine the minimum value of $\Delta v_{obj}$ several experiments
were carried out using the input image sequence with several
realizations of Gaussian white noise. Then, the minimal detectable
variation in speed, from a given reference speed $v_{obj}$, is
computed as the minimum detectable $\Delta v_{obj}$ obtained in
90\% of the experiments.

%result description
We summarize our results in Fig.~\ref{fig:discrim_OFC2_OForig}. Figure~\ref{fig:discrim_OFC2_123} shows the discrimination of the proposed parallel multi-scale algorithm for different values of L. The range of discriminated speeds is enlarged when the number of levels used in the multi-scale representation increases.
In comparison with the serial multi-scale, our method has a similar range of speed discrimination when the same number of levels are used, see Fig.~\ref{fig:discrim_OFC2_OForig_n2}, \ref{fig:discrim_OFC2_OForig_n3} and \ref{fig:discrim_OFC2_OForig_n4}. Considering both mean and variance of the discrimination in the range of speeds from 1 to 15 pixels per frame, the parallel multi-scale method shows lower values. For the case of $L=3$, parallel discrimination has mean$=14.1$ and variance$=2.2$, while serial discrimination has mean$=15.5$ and variance$=4.2$. This indicates that the proposed parallel algorithm presents a better discrimination in this range.
%"Figure 5(a) shows the discrimination of the proposed parallel
%multi-scale algorithm for different values of L. The range of
%discriminated speeds is enlarged when the number of levels used in the
%multi-scale representation increases. In comparison with the serial
%multiscale approach, our method has a similar range of speed
%discrimination when the same number of levels are used, see Figure
%5(b). Considering both the mean and the variance of the discrimination
%in the range of speeds from 1 to 16 pixels per frame, the parallel
%multi-scale method shows lower values (parallel: mean=14.1, variance
%2.2; serial:  mean=15.5, variance 4.2). This indicates that the
%proposed parallel algorithm presents a slightly better discrimination
%in this range."

\section{Discussion}
Recent works \cite{key:simon,key:chey} have
developed the idea of multi-scale estimation of speed. First,
Simoncelli \cite{key:simon} proposes a bayesian scheme
to compute the error distributions and then to estimate the velocity
using a Kalman filter through the space of scales (not time).
This approach builds a far more sophisticated error function, but it is still serial. Our work assumes that the error functions are fixed, while \cite{key:simon}assumes the error changes with respect to $\nabla I(\vec{x}, t)$, and this might be important in real-world scenarios. On the other hand, Chey et al. \cite{key:chey} propose that considering higher threshold levels
for higher scales (scale-proportional thresholds) and inter-scale
competition could explain human speed discrimination curves. We have
presented a scheme where the response of each scale regulates the
relevance of the responses of that scale. Since we handle all scales
at the same time, it corresponds to a notion of threshold and competition.
To our knowledge, no other work
models error functions as Gaussians in the log space (this
strengthen the idea that detection is not symmetrical), which seems
to fit recent recordings of motion sensitivity of neurons \cite{key:logmt}.

Finally, about the time complexity of our algorithm.
Lets consider the size of the image as $N$, the order of the optical flow algorithm as $K$ (clearly $K(N)$) and the number of scales $l$. The complexity order of the serial multi-scale algorithm is $lN + lK$, and $N + lK$ for the parallel algorithm. The only difference is in the operations involved in the merge of scales. Considering the possible speed-up using $p$ processor for the case $p=l$, then $S_p=\frac{lN + lK}{N + Kl/p}$, what can be also written as $S_p=p$. This last equation show us that the degree of parallelism (taking one level by processor) achieved by our proposed algorithm is linear.

%Perceptive experiences in humans have characterized the global
%responses in the human brain to moving stimuli, at the same time
%multiple neurons recordings in primates give information about
%theneuron population, and this gives an step forward to understand the
%human motion perception capabilities. In the other hand, living
%creatures do not require to perform sub-pixel-precise 3D
%reconstruction and it seems far fetched that bio-inspired methods will
%be a well suited solution in all motion detection related
%problems. However, bio-inspired methods, in this case were wide ranges
%of speeds can be detected, and discrimination can be performed
%(relative error), is more close to real world scenarios, and it is in
%these situation where visual bio-inspired methods such as the
%presented one show their advantages and potentiality.

\section{Conclusions} In this work we have presented a parallel
multi-scale algorithm to perform the estimation of motion using two
consecutive images. This method takes bio-inspiration from human
physiology and psychophysics knowledge in the sense that it achieves wide
uniform relative discrimination properties by using evenly spaced
logarithmic scales, and it gives results in constant time as a function
of scales. With respect to the classical serial multi-scale optical
flow algorithm, the error propagation among scales appears less
important for our proposed algorithm in terms of
relative discrimination. We now explore the idea of
using more biologically plausible methods of optical flow extraction
and the integration with a foveated topology.

\bibliographystyle{unsrt}
\bibliography{speeddiscrimination}

\end{document}